\documentclass[10pt,twocolumn,letterpaper]{article}

\usepackage[final]{cvpr}
\usepackage{times}
\usepackage{epsfig}
\usepackage{graphicx}
\usepackage{amsmath}
\usepackage{amssymb}
\usepackage{booktabs}
\usepackage{multirow}
\usepackage{paralist}
\usepackage{soul}
\usepackage{algorithm}
\usepackage[noend]{algpseudocode}
\usepackage{microtype}
\usepackage[dvipsnames, table]{xcolor}
\usepackage{enumitem}
\usepackage{bbm}
\usepackage[toc,page]{appendix}

\newenvironment{tight_itemize}{
\begin{itemize}[leftmargin=10pt]
  \setlength{\topsep}{0pt}
  \setlength{\itemsep}{2pt}
  \setlength{\parskip}{0pt}
  \setlength{\parsep}{0pt}
}{\end{itemize}}

\newcommand{\yht}[1]{{\color{red}#1}}

\usepackage[pagebackref=true,breaklinks=true,colorlinks,bookmarks=false,citecolor=blue,linkcolor=blue]{hyperref}

\begin{document}

\title{Learning Semantic Segmentation from Multiple Datasets with Label Shifts}

\author{Dongwan Kim$^1$, Yi-Hsuan Tsai$^2$\thanks{Currently at Phiar Technologies.}, Yumin Suh$^2$ \\
Masoud Faraki$^2$, Sparsh Garg$^2$, Manmohan Chandraker$^{2,3}$, Bohyung Han$^1$ \\
Seoul National University$^1$, NEC Labs America$^2$, UC San Diego$^3$ \\
}

\maketitle
\begin{abstract}
With increasing applications of semantic segmentation, numerous datasets have been proposed in the past few years. Yet labeling remains expensive, thus, it is desirable to jointly train models across aggregations of datasets to enhance data volume and diversity. However, label spaces differ across datasets and may even be in conflict with one another. This paper proposes UniSeg, an effective approach to automatically train models across multiple datasets with differing label spaces, without any manual relabeling efforts.
Specifically, we propose two losses that account for conflicting and co-occurring labels to achieve better generalization performance in unseen domains.
First, a gradient conflict in training due to mismatched label spaces is identified and a class-independent binary cross-entropy loss is proposed to alleviate such label conflicts.
Second, a loss function that considers class-relationships across datasets is proposed for a better multi-dataset training scheme.
Extensive quantitative and qualitative analyses on road-scene datasets show that UniSeg improves over multi-dataset baselines, especially on unseen datasets, e.g., achieving more than 8\% gain in IoU on KITTI averaged over all the settings.

\end{abstract}


\section{Introduction}

Recent years have seen a number of semantic segmentation datasets being proposed, for example, Cityscapes \cite{cityscapes}, BDD \cite{bdd}, IDD \cite{idd} and Mapillary \cite{mapillary}, to name a few pertaining to road scenes. These have been leveraged by semantic segmentation models to produce high-quality results \cite{deepparsing,piecewise,fcn_pami,deeplab,dilated,pspnet,Zhang_cvpr_18,crfasrnn}. However, most methods exploit labels within a single dataset for training. Given the expense of labeling segmentation data, it is important to consider whether labels from multiple datasets can be combined to train more robust models.

One benefit of a model trained on multiple datasets would be increased data volume and diversity, which might allow a joint model to better reason about challenging objects or scenes.
Moreover, if a segmentation model could be trained on a unified label space across categories, one may obtain richer training constraints and inference outputs compared to the label space of a single dataset.
However, simply combining all the datasets is non-trivial, since their label spaces are defined differently and may possibly be in conflict with one another.
For example, Cityscapes has 19 categories while Mapillary has 65 more fine-grained categories.
A recent work, MSeg \cite{mseg}, deals with this issue by manually defining a taxonomy for the unified label space, which requires re-annotating many images to achieve consistency across datasets. This makes the approach time-consuming and difficult to scale to more datasets collected in the future.

\begin{figure}[!t]
	\centering
	\includegraphics[width=\linewidth]{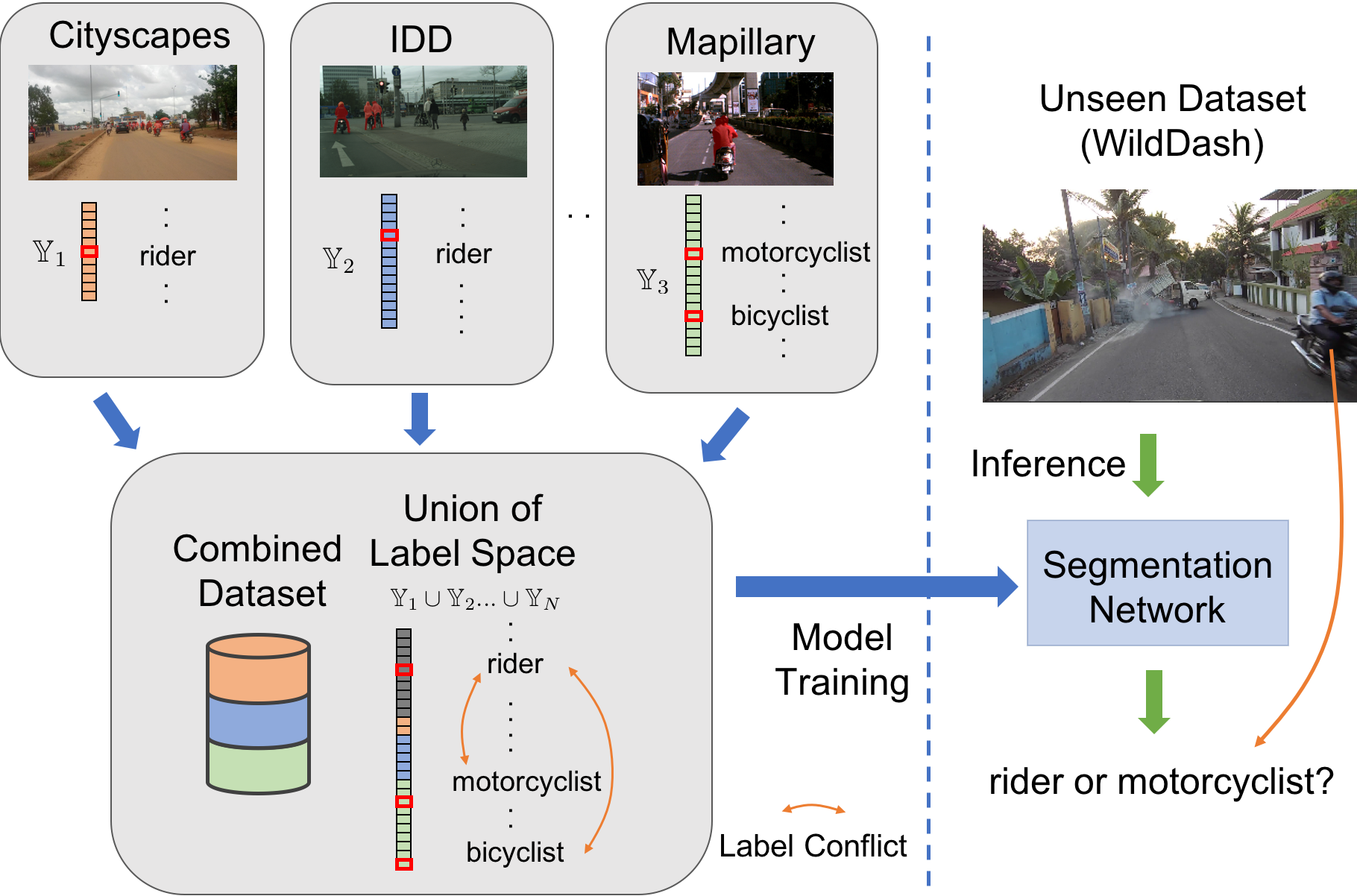}
	\caption{
	We work on the problem of multi-dataset semantic segmentation, where each dataset has a different label space. We show that directly combining all the datasets and train the model would result in a gradient conflict issue when there is a label conflict in the unified label space. Such an issue could impact the testing time when inputting an image from an unseen dataset to the model. For example, the rider in the right image can be considered as both the ``rider'' or the ``motorcyclist'' categories in the unified space. Therefore, it is of great importance to develop a method that considers such label conflict during the training process.
	}
	\label{fig:teaser}
	\vspace{-3mm}
\end{figure}

In this paper we propose a framework called \textit{UniSeg}, where we tackle the problem from the model-training perspective by designing a training scheme to automatically handle multi-dataset training with label shifts for semantic segmentation.
To this end, we propose two losses that can effectively handle conflicting labels and make use of co-occurring labels across datasets.
First, we observe that the widely-used cross-entropy loss may lead to conflicting gradients when two datasets contain categories that are in direct conflict with each other (see Figure \ref{fig:teaser}), e.g., Cityscapes only has the ``rider'' class, but Mapillary contains both the ``motorcyclist'' and ``bicyclist'' categories.
Therefore, we revisit the binary cross-entropy (BCE) loss for semantic segmentation.
Unlike the traditional cross-entropy loss, the binary cross-entropy loss enables us to compute individual gradients for each class, which resolves the gradient conflict issue during optimization on the unified label space by selectively ignoring certain classes during loss computation.
We later show that this simple modification in the loss term, which we call the \textit{Null BCE}, leads to several significant benefits for multi-dataset training, especially on unseen datasets.

To utilize the class relations across label spaces, we propose another variant of the loss, \textit{class-relational BCE}, that allows each pixel to be supervised with multiple labels.
For example, if we are able to link ``bicyclist'' from Mapillary to the Cityscapes ``rider'' class, the training process can be improved by leveraging such a relationship.
Without any external knowledge, we infer class relationships and generate multi-class labels that appropriately link categories across datasets, which are then integrated into the training process with our class-relational BCE loss.

In our experiments, we focus on road-scene datasets for semantic segmentation, which includes four training datasets: Cityscapes \cite{cityscapes}, BDD \cite{bdd}, IDD \cite{idd}, and Mapillary \cite{mapillary}. We test on three datasets, KITTI \cite{kitti}, CamVid \cite{camvid} and WildDash \cite{wilddash}, which are not used for training and serve as unseen datasets to verify that our model can perform well in other challenging conditions. 
In addition, we design a leave-one-out setting to train a model on three of the four training datasets, while the remaining dataset is left as an unseen dataset for evaluation.
We conduct extensive experiments to demonstrate the effectiveness of our \textit{UniSeg} framework with the proposed loss terms: Null BCE loss and class-relational BCE loss. We compare against baselines using the traditional multi-dataset training.
For instance, on the KITTI dataset, our HRNet-W48 \cite{hrnet} model achieves more than 8\% gain in IoU on average across all the settings. 
We also conduct qualitative analyses on the output of our UniSeg model and observe that it can make accurate multi-label predictions, especially for classes with label conflict.
In summary, the main contributions of this paper are:
\vspace{-0.2cm}
\begin{tight_itemize}
    \item We design a principled framework for multi-dataset semantic segmentation with label shifts, without introducing additional cost of human annotation or prior knowledge.
    
    \item We propose simple yet effective loss terms to handle the label shift problem via revisited BCE, while also introducing a new training scheme via class-relational BCE.

    \item We validate the benefit of using our method in various multi-dataset training settings, showing significant performance improvements over other baselines, especially on datasets that are not seen during training. Our visualizations also verify that our model is able to make more fine-grained multi-label predictions on areas with conflicting labels.
    
\end{tight_itemize}

\section{Related Work}

Since our goal is to design a framework for multi-dataset semantic segmentation, we mainly discuss the literature relevant to cross-domain settings on semantic segmentation or other scene understanding tasks, as well as domain adaptation and domain generalization problems that share a similar problem context.

\vspace{-4mm}
\paragraph{Multi-dataset Semantic Segmentation.}
A few recent works have considered to use multiple datasets to jointly train a semantic segmentation model \cite{mseg,Kalluri_iccv_19,Liang_cvpr_18,Meletis_iv_18,Bevandic_gcpr_19,Wang_cvpr21}. However, the adopted setting/goal and the perspective of their proposed approaches vary significantly.
For instance, the work of \cite{Bevandic_gcpr_19} uses Cityscapes \cite{cityscapes} as the output label space and includes the Mapillary \cite{mapillary} dataset to jointly train a semantic segmentation model and also detect outlier regions in the unseen dataset Wilddash \cite{wilddash}.
Kalluri \etal\cite{Kalluri_iccv_19} further address the problem of using two datasets in a semi-supervised learning manner, in which each dataset still retains its original label space.
Moreover, a recent work \cite{Wang_cvpr21} considers multi-dataset training using dataset-specific classifiers with the original label space of each dataset.
In contrast to our setting, these methods do not consider the unified label space using a shared classifier.

To exploit different label spaces across datasets, two approaches \cite{Liang_cvpr_18,Meletis_iv_18} adopt the idea of label hierarchy to jointly train on multiple datasets.
However, such strategy requires a manually pre-defined hierarchy label space, where categories need to be merged, added, or split in the hierarchy tree. Thus it may not be easily scalable to newly introduced datasets.
In addition, these approaches do not conduct evaluations on the unseen datasets, or only treat multi-dataset training as a pre-training method for the further fine-tuning purposes \cite{Liang_cvpr_18}.
More recently, MSeg \cite{mseg} proposes to unify multiple datasets via defining a label taxonomy, which maps all the datasets into the same label space.
However, this pre-processing scheme requires human re-annotation on a large number of images to ensure label consistency, which is time-consuming and not scalable to more datasets.
In contrast, we aim to tackle the multi-dataset setting with label shifts, purely from the model-training perspective without involving any manual process.

\vspace{-4mm}
\paragraph{Multi-dataset Scene Understanding.}

In addition to semantic segmentation, a few efforts on multi-dataset object detection \cite{Wang_UniDet_cvpr19,Zhao_UniDet_ECCV20,Xu_aaai_20,Liu_arxiv_20,Zhou_unidet} and depth estimation \cite{midas} have been made.
Wang \etal \cite{Wang_UniDet_cvpr19} first propose to train a single detector from multiple datasets in a multi-task fashion, in which each dataset has its own prediction head to obtain individual outputs,
while \cite{Zhou_unidet} improves the usage of individual detection heads by optimizing the mapping from each dataset-specific label to one common label space.
The work of \cite{Zhao_UniDet_ECCV20} further operates on the setting of unifying all the label spaces, and then utilizes a pseudo-label re-training scheme to recover the missing labels in each dataset.
Moreover, \cite{Liu_arxiv_20} tackles cross-dataset object detection in two label spaces under an incremental learning setting, where a distillation-based method is applied to reduce the feature gap.
In this work, our focus is to tackle the label-shift and conflict issues in multi-dataset semantic segmentation using one shared classifier on the unified label space, which has a different characteristic from the aforementioned object detection and depth estimation tasks.

\vspace{-4mm}
\paragraph{Domain Adaptation and Generalization.}
Unsupervised domain adaptation (DA) techniques have been developed to learn domain-invariant features that reduce the gap across the source and target domains in  several computer vision tasks, such as image classification \cite{long2015learning,ganin2016domain,tzeng2017adversarial,Saito_CVPR_2018,Lee_DTA_ICCV19}, object detection \cite{YChenCVPR18,SaitoCVPR19,KimCVPR19,Hsu_uda_det_eccv20}, and semantic segmentation \cite{Hoffman_ICML_2018,Tsai_DA4Seg_ICCV19,Vu_CVPR_2019,Li_CVPR_2019,Paul_daseg_eccv20}.
Extending from a traditional dual-domains setting, other DA settings have been proposed to consider multi-source \cite{domainnet} and multi-target \cite{Liu_cvpr_20,dai2019adaptation} domains.
Furthermore, a more challenging setting, universal DA \cite{Saito_nips_20,You_cvpr_19}, deals with various cases across datasets that may have different label spaces.
In our multi-dataset setting with supervisions, although there are also domain gaps across datasets similar to the domain adaptation setting, we focus on solving the label shift and conflict issues, which is orthogonal to the adaptation scenario.

Domain generalization assumes multiple training datasets available, and the goal is to learn a model that can generalize well to unseen datasets.
Several methods have developed via learning a share embedding space \cite{Muandet_icml13,Dou_nips19,Motiian_iccv17,Faraki_cvpr21}, domain-specific information \cite{Seo_eccv20,Li_iccv17}, or meta-learning based approaches \cite{Balaji_nips19,Li_aaai19}.
However, these approaches mainly focus on the image classification task, and more importantly, assume a shared label space across the training datasets and any unseen ones, which is different from the setting of this paper, where each dataset may have its own distinct label space.

\begin{figure*}[t]
    \centering
    \includegraphics[width=0.95\textwidth]{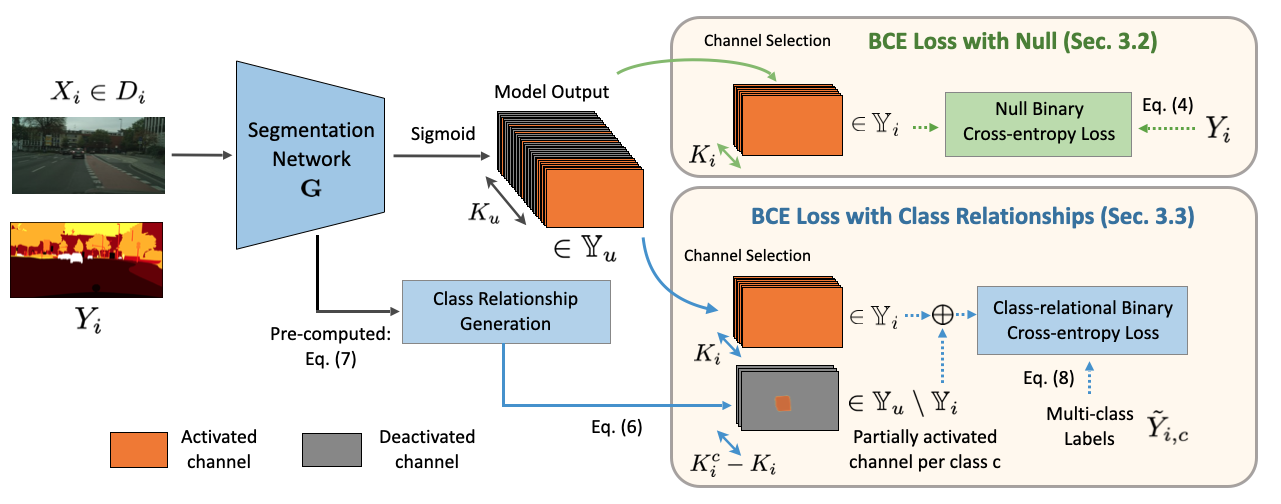}
    \vspace{-3mm}
    \caption{
    Overview of the proposed framework using the proposed Null BCE loss (Section \ref{sec:bce}) and Class-relational BCE loss (Section \ref{sec:bce-cl}): 1) For Null BCE loss, we replace the original cross-entropy loss function to reduce the gradient conflict issue as mentioned in Section \ref{sec:ce}. Specifically, in the loss calculation, we only take the categories in the label space $\mathbb{Y}_i$ of $D_i$ into consideration via \eqref{eq:loss_bce}; 2) For class-relational BCE loss, through the pre-computed class relationships for each dataset via \eqref{eq:class_distribution}, we incorporate the generated multi-class labels $\tilde{Y}_{i,c}$ via \eqref{eq:multi-class} into the Null BCE loss and form a class-relational BCE loss via \eqref{eq:loss_class} to better enable training across multiple datasets. Note that the multi-class label may only contain partial pixels within the image as illustrated (denoted as the orange color).
    }
    \label{fig:overall_pipeline}
    \vspace{-3mm}
\end{figure*}

\section{Proposed Method}

\subsection{Multi-dataset Semantic Segmentation}
\label{sec:ce}
The typical way to optimize a single-dataset semantic segmentation model is to use a pixel-wise cross-entropy loss. When it comes to multiple datasets, since each dataset has its own label space, there could be two straightforward options to train the model. One method is to construct individually separate classifiers for each dataset. 
However, this could result in problems during testing, where it may not be clear which classifier should be selected.
The second option is to unify all the label spaces from multiple datasets and train a single classifier, which is more suitable for our problem context and will be the strategy on which this paper focuses.

\vspace{-4mm}
\paragraph{Cross-Entropy Formulation.}
Given an image $X_i \in \mathbb{R}^{H \times W \times 3}$ in dataset $D_i$ and its $K_i$-categorical one-hot label $Y_i \in \{0,1\}^{H \times W \times K_i}$ in the label space $\mathbb{Y}_i$, we unify the label space as $\mathbb{Y}_{u} = \mathbb{Y}_1 \cup \mathbb{Y}_2 ...\cup \mathbb{Y}_N$, where $N$ is the number of datasets.
Therefore, the original label $Y_i$ is extended to $K_u$ categories, where $K_u \leq \sum_{i} K_i$ is the number of unified categories.
Without any prior knowledge, a natural way to extend $Y_i$ to a $K_u$-categorical label is to assign all categories in $\mathbb{Y} = \mathbb{Y}_u \setminus \mathbb{Y}_i$ with label 0.
As a result, the cross-entropy loss that optimizes the segmentation network $\mathbf{G}$ on multiple datasets can be written as:
\begin{equation}
	\mathcal{L}_{seg}^{ce} =
	- \sum_{i=1}^{N} \sum_{k=1}^{K_u} \sum_{h,w}{}  Y_i^{(h, w, k)} \log(P_i^{(h,w,k)}) \;,
	\label{eq:loss_ce}
\end{equation}
where $P_i \in [0, 1]^{H \times W \times K_u}$ is the softmax of the segmentation output $O_i = \mathbf{G}(X_i) \in \mathbb{R}^{H \times W \times K_u}$, from the unified classifier. In \eqref{eq:loss_ce}, we omit summation over all samples in each dataset to prevent notations from being over-complicated. 

\vspace{-4mm}
\paragraph{Gradient Conflict in \eqref{eq:loss_ce}.}
Although unifying the label spaces across datasets enables the standard cross-entropy optimization in \eqref{eq:loss_ce}, it can cause training difficulty when there is a label conflict across datasets.
Here, we assume that we do not have prior knowledge regarding the label space and its semantics in the individual datasets. Therefore, such label conflict is likely to occur, as each dataset may define label spaces differently. For instance, Cityscapes only has the ``rider'' class, while Mapillary does not, and has the ``motorcyclist'' and ``bicyclist'' categories instead. In this case, the unified label space of $\mathbb{Y}_u$ contains all three categories, ``rider'', ``motorcyclist'', and ``bicyclist'', but during training, images from Mapillary would always treat ``rider'' with a label of 0.

Based on the example above, such label conflict may cause optimization difficulty with the cross-entropy (CE) loss, especially due to the fact that the softmax function is dependent on the outputs of all classes. To further analyze the negative effect caused by label conflict, we consider an example and show the step of updating a single parameter $\theta$ that contributes to the output $O$ of an arbitrary class $k$ in the last layer of the network. Given an image $X_1$ from one dataset that is labeled as $k$ at position $(h,w)$, the gradient propagated by the loss at position $(h, w)$ to parameter $\theta$ can be calculated as:
\begin{align}
    \frac{ \partial \mathcal{L}_{seg}^{ce} } { \partial \theta } = 
    \frac{\partial O_1^{(h,w,k)}}{\partial \theta} (P_1^{(h,w,k)} - Y_1^{(h, w, k)}).
    \label{eq:conflict1}
\end{align}
Now, consider an identical image $X_2$ that originates from another dataset defined by a different label space. Combining the two cases, the gradient update for parameter $\theta$ becomes:
\begin{align}
    \frac{ \partial \mathcal{L}_{seg}^{ce} } { \partial \theta } = &  
     \frac{\partial O_1^{(h,w,k)}}{\partial \theta} (P_1^{(h,w,k)} - Y_1^{(h, w, k)}) \notag\\ 
    &+ \frac{\partial O_2^{(h,w,k)}}{\partial \theta} (P_2^{(h,w,k)} - Y_2^{(h, w, k)}).
    \label{eq:conflict}
\end{align}
Note that, since $X_1$ and $X_2$ are identical images, $\frac{\partial O_1^{(h,w,k)}}{\partial \theta} = \frac{\partial O_2^{(h,w,k)}}{\partial \theta}$. However, since the two images originate from different datasets, we have $Y_1^{(h, w, k)} \neq Y_2^{(h, w, k)}$ (\ie, if $Y_1^{(h, w, k)}=1$, $Y_2^{(h, w, k)}=0$). Thus, the parameter $\theta$ receives one gradient that is smaller than 0, and another that is larger than 0, despite coming from identical samples. This is not optimal for training the model, yet can easily occur when training a model on multiple datasets with conflicting label spaces.

\subsection{Revisited Binary Cross-Entropy Loss}
\label{sec:bce}

To resolve the aforementioned issue, we find that the binary cross-entropy (BCE) loss, while similar to the CE loss, exhibits some interesting properties that make it favorable for our task setting. First, it does not require a softmax operation, whose value is dependent on the output logits of other classes. Instead, BCE loss is accompanied by a Sigmoid activation on the outputs, which is independently applied to each class. Furthermore, with the BCE loss, we are able to selectively assign labels to each class. Therefore, we design a ``Null'' class strategy, where we only assign the \textit{valid} labels for each dataset.
That is, for images from the $D_i$ dataset, we only assign labels for categories within $\mathbb{Y}_i$ as it is, while for other categories $\mathbb{Y} = \mathbb{Y}_u \setminus \mathbb{Y}_i$, we neither assign label 0 nor 1. We name this loss as the ``Null BCE loss''.
As a result, our Null BCE loss can be written as:
\begin{align}
	\mathcal{L}_{seg}^{bce} =
	- \sum_{i=1}^{N} \sum_{k=1}^{\textcolor{blue}{K_i}} \sum_{h,w} Y_i^{(h,w,k)} \log(Q_i^{(h,w,k)}) \notag \\ 
	+ (1 - Y_i^{(h,w,k)}) \log(1-Q_i^{(h,w,k)}) \;,
	\label{eq:loss_bce}
\end{align}
where $Q_i \in [0,1]^{H \times W \times K_u}$ is the output from the Sigmoid activation that represents the independent probability of each class. 
It is important to note that, although there is only a slight difference from \eqref{eq:loss_ce} in the summation of the loss term (i.e., summed over class $K_i$ here, highlighted in blue), this change makes a difference in gradient updates, leading the gradient conflict issue mentioned in \eqref{eq:conflict} to be resolved, since no loss is calculated for class $k$ given the input image $X_2$ (see the example in \eqref{eq:conflict}):
\begin{equation}
    \frac{ \partial \mathcal{L}_{seg}^{bce} } { \partial \theta } =  
    \frac{\partial O_1^{(h,w,k)}}{\partial \theta} (Q_1^{(h,w,k)} - Y_1^{(h, w, k)}).
    \label{eq:bce_grad}
\end{equation}
The procedure is illustrated in the top-right of Figure \ref{fig:overall_pipeline}.
A more detailed derivation for both \eqref{eq:conflict1} and \eqref{eq:bce_grad} is provided in the supplementary material.

\subsection{Class-relational Binary Cross-Entropy Loss}
\label{sec:bce-cl}

Another advantage of BCE over the CE loss is that it can be used to train a model with multi-label supervision. 
While our Null BCE loss in Section \ref{sec:bce} alleviates the gradient conflict issue caused by inconsistent label spaces across multiple datasets, it simply chooses to ignore classes that are not within the label space of a given sample.
Thus, we propose another loss that better utilizes the inter-class relationships, by explicitly providing multi-label supervision at pixels where co-occurring labels exist (bottom-right of Figure \ref{fig:overall_pipeline}).

\vspace{-4mm}
\paragraph{Multi-class Label Generation.}
For a class $c$ from dataset $D_i$, we generate the new multi-class label $\tilde{Y}_{i,c} \in \{0,1\}^{K_u}$. This aims at training the classifier to predict not only the original class but also the co-existing class from the unified label space $\mathbb{Y}_u$ when they are semantically related.
For example, such multi-class labels can be generated for subset/superset relationships, \eg, ``bicyclist'' co-existing with ``rider'' or ``crosswalk'' co-existing with ``road'', but not classes with similar appearance or high co-occurrence.

Assuming we know these class relationships, we assign additional label $c' \in \mathbb{Y}_u$ only if its similarity to the class $c \in \mathbb{Y}_i$ is sufficiently large,
\begin{equation}
    \tilde{Y}_{i,c}^{(c')} = 
    \begin{cases}
      1 & \text{if } c' = c \text{ or } \mathbf{s}_{i,c}^{(c')} > \max(\tau, \mathbf{s}_{i,c}^{(c)}) \\
      \O & \text{else if } c' \in \mathbb{Y}_u \setminus \mathbb{Y}_i \\
      0 & \text{else},
    \end{cases}
    \label{eq:multi-class}
\end{equation}
where $\mathbf{s}_{i,c}^{(c')}$ is the similarity between class $c$ and $c'$ measured in dataset $D_i$ (details for calculation are introduced in the next section) and $\tau$ is a threshold.
When the class $c$ and $c'$ have a conflict, \eg when $c$ is ``bicyclist'' and $c'$ is ``rider'', we expect the similarity $\mathbf{s}_{i,c}^{(c')}$ to be large. In contrast, we expect it to be small for classes without conflict.

\begin{figure}[t]
	\centering
	\footnotesize
	\setlength{\tabcolsep}{2pt}
	\includegraphics[width=0.95\columnwidth]{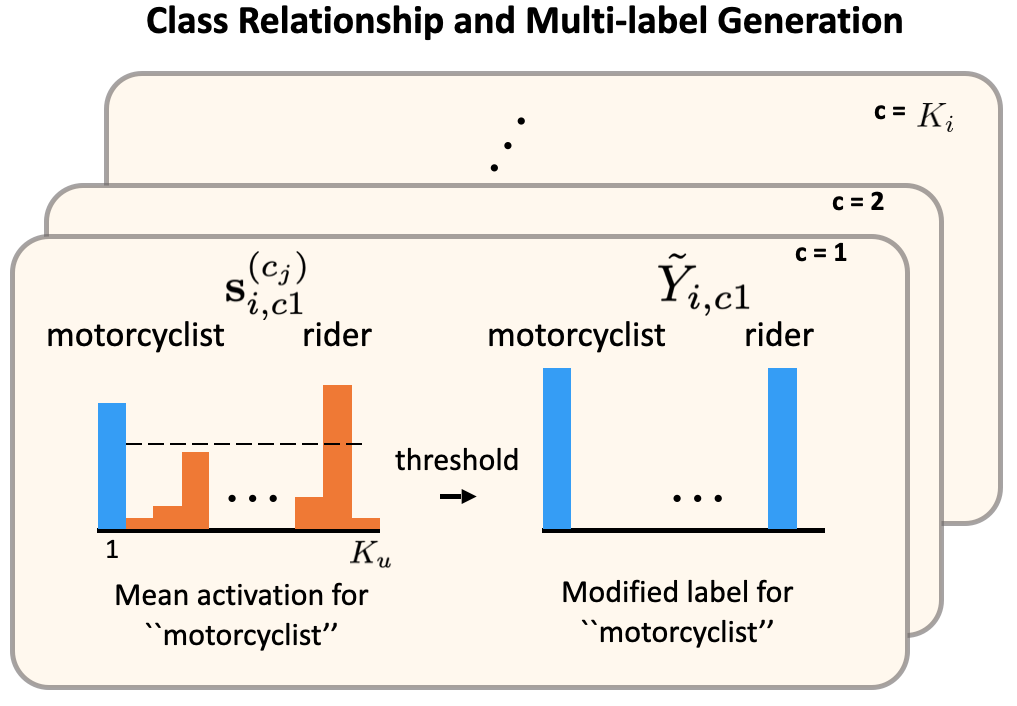}
	\vspace{-3mm}
	\caption{
	One example of generating the final multi-class label $\tilde{Y}_{i, c}$ through the mean activation $\mathbf{s}_{i, c}^{(c)}$ for ``motorcyclist'', where the final multi-class includes the ``rider'' class.
	}
	\label{fig:class_relation}
	\vspace{-5mm}
\end{figure}

For the choice of $\tau$, we design an automatic way by checking if the class of the largest score in $\mathbf{s}_{i,c}$ comes from another dataset, $D_j$, which indicates that there is likely a label conflict and requires multi-class labels. For such cases, we average the largest scores and obtain a value of 0.48$\pm$0.01, which is used as the threshold $\tau$.
Note that, the $\max$ condition in \eqref{eq:multi-class} implies that multi-labels are only activated, \ie, $\tilde{Y}_{i,c}^{(c')}=1$, when similarity for class $c' \in \mathbb{Y}_u \setminus \mathbb{Y}_i$ is higher than that of the original class $c$, \ie, $\mathbf{s}_{i,c}^{(c')} \geq \mathbf{s}_{i,c}^{(c)}$.
This makes label generation more robust to variations in $\tau$.
One illustration of this process is shown in Figure \ref{fig:class_relation}.

\vspace{-4mm}
\paragraph{Class Relationship Generation.}
To extract inter-class relationships, we leverage the cosine classifier \cite{cosface}, such that the cosine similarity between the feature and any classifier weight vector can be calculated, even for label spaces across datasets.
More details for cosine classifier are provided in the supplementary material.
With the cosine similarity, we then calculate the mean activation vector of the final output layer as the similarity score, $\mathbf{s}_{i, c} \in [0, 1]^{K_u}$, which indicates the relationships between each class $c$ in dataset $D_i$ and all other classes in the unified label space $\mathbb{Y}_u$. 
\begin{align}
    \label{eq:class_distribution}
    \mathbf{s}_{i, c}^{(c')} = \frac{1}{M_{i, c}}\sum_{X_i \in D_i}\sum_{h,w} S_i^{(h,w,c')} \cdot \mathbbm{1}_{i,c}^{(h, w)};\notag\\ 
    \forall i \in \{1, ..., N\}; \forall c' \in \mathbb{Y}_u,
\end{align}
where $S_i^{(h,w,c')}$ is the cosine similarity between the image feature from dataset $D_i$ and the weight of class $c'$, $\mathbbm{1}_{i,c}^{(h,w)} \in \{0, 1\}$ is an indicator whose value is 1 if the ground-truth is $c$ at location $(h, w)$ of $X_i$.
$M_{i,c}$ denotes the number of pixels with ground-truth of $c$ in $D_i$, and $X_i$ represents the samples in $D_i$.
Note that, $\mathbf{s}_{i, c}$ can be computed either for each dataset or over all the datasets. In practice, although we do not find much empirical difference between these two schemes, we adopt the dataset-specific similarities to reflect the properties of each individual dataset.

\vspace{-4mm}
\paragraph{Discussions.}
In \eqref{eq:class_distribution}, we define the similarity between classes asymmetric, \ie $s_{i,c}^{c'} \neq s_{j,c'}^{c}$, where $i\neq j$ and $c' \in \mathbb{Y}_j$, in order to address the asymmetric relations such as subset/superset.
For example, there is a label conflict between ``rider'' in Cityscapes and ``motorcyclist'' in Mapillary. Since ``rider'' is a superset of ``motorcyclist'', any ``motorcyclist'' is also a ``rider'', yet the opposite is not always true.
Here, our method is able to implicitly capture such intricate relationships, where the model can generate stronger ``rider'' activations given inputs of ``motorcyclist''.
On the contrary, the model does not generate strong ``motorcyclist'' activations on ``rider'', since a ``rider'' is not always a ``motorcyclist''.
Therefore, we incorporate these relationships into our training in the class-relational BCE loss.

\vspace{-4mm}
\paragraph{Class-relational BCE Loss.}
With the the multi-class label $\tilde{Y}_{i, c}$ via \eqref{eq:multi-class} that is aware of the class-relationships across datasets, we define our class-relational BCE Loss as:
\begin{align}
	\mathcal{L}_{seg}^{cl-bce} =
	- \sum_{i=1}^{N} \sum_{k=1}^{\textcolor{blue}{K_i^c}} \sum_{h,w} \textcolor{blue}{\tilde{Y}_{i,c}^{(h,w,k)}} \log(Q_i^{(h,w,k)}) \notag \\ 
	+ (1 - \textcolor{blue}{\tilde{Y}_{i,c}^{(h,w,k)}}) \log(1-Q_i^{(h,w,k)}) \;,
	\label{eq:loss_class}
\end{align}
where the difference (denoted in blue above) from \eqref{eq:loss_bce} is the summation over the $K_i^c$-categorical multi-label $\tilde{Y}_{i,c}$ that is calculated for each class $c$.
As a result, some of the ``Null'' categories from Section \ref{sec:bce} can now be incorporated in the loss calculation based on the inferred class relationships.
The full list of generated multi-labels are provided in the supplementary material.

\subsection{Model Training and Implementation Details}

\paragraph{Data Preparation.}
As noted in Section~\ref{sec:ce}, given $N$ number of datasets, $D = \{D_1, D_2,...,D_N\}$, we unify the label space as the union of the $N$ individual label spaces, $\mathbb{Y}_{u} = \mathbb{Y}_1 \cup \mathbb{Y}_2 ...\cup \mathbb{Y}_N$. 
We combine the $N$ datasets by a simple concatenation operator to obtain a single unified dataset.
Before doing so, we preprocess each dataset such that the segmentation labels can be re-mapped to the correct index of $\mathbb{Y}_{u}$. Note that, to make the training batches consistent, we resize all the images with the shorter side as 1080 pixels, and use $713 \times 713$ random cropping with standard data augmentations such as scaling and random horizontal flipping.

\vspace{-4mm}
\paragraph{Implementation Details.}
We use the HRNet V2~\cite{hrnet} backbones initialized with weights pre-trained on ImageNet~\cite{imagenet}. The batch size is 16/32 with an initial learning rate of 0.01/0.02 for HRNet-W18 and HRNet-W48, respectively. All models are trained using SGD with momentum and a polynomial learning rate decay scheme.
To obtain the multi-class labels in our class-relational BCE loss, we pre-train a model using the cosine classifier and fix the generated class relationships $\mathbf{s}_{i, c}$ for each dataset in all the experiments.

\section{Experimental Results}

In this section, we first introduce our experimental setting on multi-dataset semantic segmentation. Then, to verify the robustness of our UniSeg model, we present the results trained using different combinations of four driving-scene datasets (Cityscapes \cite{cityscapes}, BDD \cite{bdd}, IDD \cite{idd}, and Mapillary \cite{mapillary}), and tested on three unseen datasets (KITTI \cite{kitti}, CamVid \cite{camvid} and WildDash \cite{wilddash}).
In addition, we diversify our experiments by training models on ``Leave-One-Out'' settings, where one of the four training datasets acts as a held-out testing set (unseen), and the model is trained on the remaining three datasets.
Note that, since we focus on road-scene label spaces, other disparate labels may not be handled in our case and we leave it as the future direction, \eg, incorporating indoor-type datasets.
Our quantitative results are accompanied by qualitative results, which provide more insight of our model.

\subsection{Datasets and Experimental Setting}

Here, we describe individual datasets and their dataset-specific characteristics that could affect multi-dataset training on semantic segmentation. In experiments, we use official splits in training and evaluation.

The \textbf{Cityscapes} and \textbf{BDD} datasets are collected from different environments (central Europe and USA, respectively), but both contain the same 19 classes in their label spaces. The \textbf{IDD} dataset is collected in India, and provides a hierarchical label space with four levels. We follow a conventional level-3 setting which contains 17 stuff and 9 object classes. 
Finally, the \textbf{Mapillary} dataset is one of the largest driving scenes dataset, with data collected from around the world, and has a total of 65 fine-grained categories. Overall, the unified label space when training on all four datasets has a total of 70 categories.

\textbf{KITTI}, \textbf{WildDash}, and \textbf{CamVid} are all relatively small-scale segmentation datasets that we use as unseen test datasets. The label spaces of KITTI and WildDash are identical to Cityscapes, while for CamVid, we follow the reduced label space used in~\cite{mseg}.

\vspace{-4mm}
\paragraph{Evaluation.}
To perform evaluation on each dataset that may have a different label space from the model, we select appropriate channels from the model output followed by the argmax operation.
That is, we evaluate on the classes that exist both on the label set where the model is trained from and the label set defined in the test dataset.
During testing, following \cite{mseg}, all images are resized so that the height of the image is 1080p (while maintaining the aspect ratio). 
Intersection-over-union (IoU) score is used to evaluate the final segmentation output. 

Note that, while MSeg~\cite{mseg} uses a similar set of training and evaluation datasets, our models are unable to directly compare with theirs. This is because MSeg manually defines a label taxonomy, and in the process, it merges or splits (and relabels) certain categories. For example, MSeg merges the ``lane marking'', ``crosswalk'', and many more into a single ``road'' category, which reduces the total number of classes. Since the mean-IoU (mIoU) metric is heavily dependent on the number of classes, it is hard to make direct comparisons with Mseg.

\begin{table}[!t]
	\caption{
		Mean Intersection-over-Union (mIoU) comparisons with baselines on various multi-dataset settings using the HRNet-W18 architecture. KITTI, WildDash, and CamVid are fixed as unseen test datasets. ``N/A'' indicates the setting where there are no multi-labels generated for our final model and thus we only show our Null BCE setting. ``C-R BCE'' indicates class-relational BCE.
	}
	\vspace{-2mm}
	\label{table:hrnet_small}
	\small
	\centering
	\renewcommand{\arraystretch}{1.1}
	\setlength{\tabcolsep}{3pt}
	\resizebox{0.47\textwidth}{!}{
	\begin{tabular}{llcccc}
		\toprule
		Train Datasets & Method & KITTI & WildDash & CamVid & Mean \\
		\midrule
		Single-dataset & Single-best (CE) & 41.9 & 41.2 & 60.8 & 48.0 \\
		\midrule
		\multirow{3}{*}{\shortstack[l]{C + I + B}} & Multi-dataset (CE) & 46.2 & 44.3 & 70.3 & 51.6 \\ 
		                                        & UniSeg (Null BCE) & \textbf{54.9} & \textbf{44.6} & \textbf{71.8} & \textbf{54.9} \\
		                                        & UniSeg (C-R BCE) & N/A & N/A & N/A & N/A \\
		\midrule
		\multirow{3}{*}{\shortstack[l]{I + B + M}} & Multi-dataset (CE) & 48.0 & 47.5 & 71.2 & 55.4\\
		                                        
		                                        & UniSeg (Null BCE) & 52.7 & 47.7 & \textbf{73.2} & 57.9\\
		                                        & UniSeg (C-R BCE) & \textbf{54.6} & \textbf{48.3} & 72.7 & \textbf{58.5}   \\
		\midrule
		\multirow{3}{*}{\shortstack[l]{C + I + M}} & Multi-dataset (CE) & 50.2 & 41.3 & 72.8 & 54.8\\
		                                        & UniSeg (Null BCE) & 55.9 & 44.1 & 73.3 & 57.8 \\
		                                        & UniSeg (C-R BCE) & \textbf{56.5} & \textbf{44.8} & \textbf{73.8} & \textbf{58.4}\\
		\midrule
		\multirow{3}{*}{\shortstack[l]{C + B + M}} & Multi-dataset (CE) & 55.0 & 46.2 & 73.4 & 58.2\\
		                                        
		                                        & UniSeg (Null BCE) & 59.0 & 47.5 & 73.8 & 60.1 \\
		                                        & UniSeg (C-R BCE) & \textbf{59.2} & \textbf{48.7} & \textbf{74.0} & \textbf{60.6} \\ 
		\midrule
		\multirow{3}{*}{\shortstack[l]{C + I + B + M \\(All)}} & Multi-dataset (CE) & 48.8 & 46.0 & 72.7 & 55.8 \\
		                                        & UniSeg (Null BCE) & 57.6 & 47.5 & 73.3 & 59.5 \\
		                                        & UniSeg (C-R BCE) & \textbf{58.9} & \textbf{48.2} & \textbf{73.9} & \textbf{60.3}\\
		\bottomrule
	\end{tabular}
	}
	\vspace{-5mm}
\end{table}

\subsection{Overall Performance}

We present our quantitative results for the unseen datasets in Table \ref{table:hrnet_small} and \ref{table:hrnet_big} and the leave-one-out setting in Table \ref{table:leave-one-out}. For more insightful comparisons, we focus on the evaluation of unseen datasets as it is a more interesting setting for validating the generalizability of models, and leave the results for seen datasets in the supplementary material.

\vspace{-4mm}
\paragraph{Full Setting.}
In Table \ref{table:hrnet_small} and \ref{table:hrnet_big}, we show performance on unseen datasets (KITTI, WildDash, CamVid) by using the leave-one-out settings (\ie, three training datasets as in Table \ref{table:leave-one-out}) or all four training datasets to train the model.
In addition to the three methods, we present the ``single-best'' baseline, where the results are obtained by the best model that trains on each of the training datasets using the cross-entropy loss. That is, we evaluate all single-dataset models and report results of the highest performing model on unseen datasets.

\begin{table}[!th]
	\caption{
		Mean Intersection-over-Union (mIoU) comparisons with baselines on various multi-dataset settings using the HRNet-W48 architecture. KITTI, WildDash, and CamVid are fixed as unseen test datasets. ``N/A'' indicates the setting where there are no multi-labels generated for our final model and thus we only show our Null BCE setting. ``C-R BCE'' indicates class-relational BCE.
	}
	\vspace{-2mm}
	\label{table:hrnet_big}
	\small
	\centering
	\renewcommand{\arraystretch}{1.1}
	\setlength{\tabcolsep}{3pt}
	\resizebox{0.47\textwidth}{!}{
	\begin{tabular}{llcccc}
		\toprule
		Train Datasets & Method & KITTI & WildDash & CamVid & Mean \\
		\midrule
		Single-dataset & Single-best (CE) & 48.1 & 48.8 & 73.6 & 56.8 \\
		\midrule
		\multirow{3}{*}{\shortstack[l]{C + I + B}} & Multi-dataset (CE) & 54.5 & 51.5 & 76.7 & 60.9 \\ 
		                                        & UniSeg (Null BCE) & \textbf{62.9} & \textbf{54.2} & \textbf{76.8} & \textbf{64.6} \\
		                                        & UniSeg (C-R BCE) & N/A & N/A & N/A & N/A \\
		\midrule
		\multirow{3}{*}{\shortstack[l]{I + B + M}} & Multi-dataset (CE) & 54.4 & 53.5 & 77.7 & 61.9 \\
		                                        & UniSeg (Null BCE) & 58.4 & \textbf{55.9} & 77.8 & 64.0 \\
		                                        & UniSeg (C-R BCE) & \textbf{59.3} & 55.7 & \textbf{78.3} & \textbf{64.4} \\
		\midrule
		\multirow{3}{*}{\shortstack[l]{C + I + M}} & Multi-dataset (CE) & 56.8 & 52.9 & 78.0 & 62.6 \\
		                                        & UniSeg (Null BCE) &  63.9 & 53.5 & 78.4 & 65.3 \\
		                                        & UniSeg (C-R BCE) & \textbf{65.8} & \textbf{53.6} & \textbf{78.7} & \textbf{66.0} \\
		\midrule
		\multirow{3}{*}{\shortstack[l]{C + B + M}} & Multi-dataset (CE) & 57.2 & 54.2 & 78.2 & 63.2 \\
		                                        & UniSeg (Null BCE) & 64.7 & 55.3 & 78.1 & 66.0 \\
		                                        & UniSeg (C-R BCE) & \textbf{68.0} & \textbf{57.8} & \textbf{78.3} & \textbf{68.0} \\ 
		\midrule
		\multirow{3}{*}{\shortstack[l]{C + I + B + M \\(All)}} & Multi-dataset (CE) & 57.0 & 56.0 & 78.1 & 63.7 \\
		                                        & UniSeg (Null BCE) & 64.4 & 56.5 & 78.2 & 66.4 \\
		                                        & UniSeg (C-R BCE) & \textbf{65.2} & \textbf{58.4} & \textbf{78.6} & \textbf{67.4} \\
		\bottomrule
	\end{tabular}
	}
\end{table}

\vspace{-4mm}
\paragraph{Leave-one-out.}
We employ leave-one-out settings with the training datasets to diversify our testing scenarios.
In these settings, one of the four training datasets (\ie, Cityscapes, IDD, BDD, Mapillary) is left out of training and treated as an unseen test dataset.
For example, in Table \ref{table:leave-one-out}, each column presents results of the unseen test dataset, while the other three serve as train datasets.
For each setting we compare three different models: 1) using the traditional cross-entropy (CE) loss, 2) using our Null BCE loss, and 3) using our class-relational BCE loss, as well as two backbone architectures, HRNet-W18 and HRNet-W48.
Note that, when training on ``Cityscapes, IDD, BDD'', the unified label space is small and thus there are no new labels generated by~\eqref{eq:multi-class} (\ie, column ``Mapillary'' of Table \ref{table:leave-one-out}).  

\vspace{-4mm}
\paragraph{Results.}

First, we observe that jointly training on multiple datasets generally outperforms the single-best setting, even when the CE loss is used.
This shows the advantage of using multi-dataset training, in which the data diversity and volume are increased.
We observe that the two variants of our UniSeg models --- Null BCE and class-relational BCE --- consistently perform favorably against the multi-dataset baseline with the typical CE loss.
For example, using either the HRNet-W18/W-48 model, averaged over all the settings, there is a 7.2\%/8.2\% gain on KITTI and a 3.3\%/3.6\% gain on ``Mean'' in Table \ref{table:hrnet_small} and \ref{table:hrnet_big}, which is considered as a significant improvement in semantic segmentation.
Furthermore, the results in Table 3 follow a similar trend to Tables~\ref{table:hrnet_small} and~\ref{table:hrnet_big}, where the UniSeg models consistently outperform the CE baseline.
This validates our original intuition that the gradient conflict in the CE loss affects model's robustness.

Finally, comparing between our two model variants, class-relational BCE further improves the overall performance. This shows that providing our generated multi-class labels helps multi-dataset training with conflicting label spaces.

\begin{figure*}[!t]
    \vspace{-4mm}
    \centering
    \begin{tabular}{cccc}
        \multicolumn{4}{c}{\includegraphics[width=0.92\textwidth]{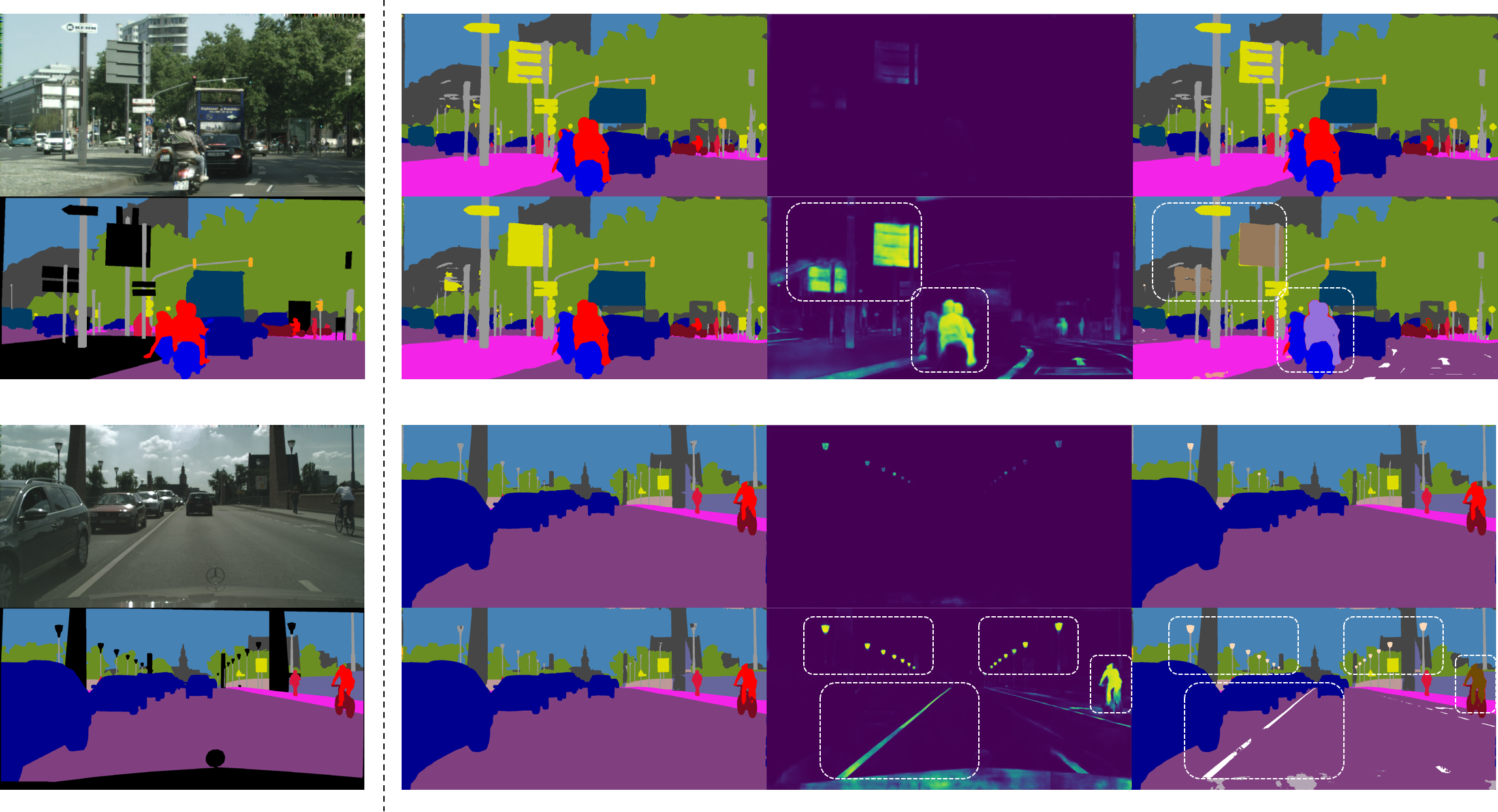}} \\
        \hspace{1cm}\shortstack{\small{Input image} \& \\ \small{ground truth}} & \hspace{1.6cm}\shortstack{\small{Prediction on} \\ \small{Cityscapes label space}} & \hspace{0.4cm}\shortstack{\small{Normalized top1 activation} \\ \small{for non-Cityscapes classes}} & \shortstack{\small{Multi-label prediction:} \\ \small{entire label space}}
    \end{tabular}
    \vspace{-3mm}
    \caption{Multi-label predictions on two samples of Cityscapes. For each sample, the first row corresponds to an HRNet-W48 model trained with the CE loss, while the second row corresponds to our C-R BCE model. While both models make strong predictions on the Cityscapes label space (column 2), only the C-R BCE model has high (normalized) activations for non-Cityscapes classes in regions with label conflict (column 3). For example, the C-R BCE model correctly predicts ``traffic sign - back'' (color: light brown) and ``traffic light'' (color: beige), even though it is not included in the ground-truth for Cityscapes. Furthermore, the C-R BCE model can make more fine-grained predictions, such as ``rider'' $\rightarrow$ ``motorcyclist'' (color: light purple), ``rider'' $\rightarrow$ ``bicyclist'' (color: brown), and ``road'' $\rightarrow$ ``lane marking'' (color: white).}
    \label{fig:visualization}
    \vspace{-5mm}
\end{figure*}

\begin{table}[!t]
	\caption{
		mIoU comparisons for the Leave-One-Out settings with HRNet-W18 and HRNet-W48. Each column indicates the unseen testing dataset, while the model is trained on the remaining three datasets. ``C-R BCE'' indicates class-relational BCE.
	}
	\label{table:leave-one-out}
	\small
	\centering
	\renewcommand{\arraystretch}{1.1}
	\setlength{\tabcolsep}{2.5pt}
	\resizebox{0.47\textwidth}{!}{
	\begin{tabular}{llccccc}
		\toprule
		Method & Arch. & Cityscapes & IDD & BDD & Mapillary & Mean \\
		\midrule
		Multi-dataset (CE) & \multirow{3}{*}{\shortstack{HRNet\\W18}} & 55.0 & 44.8 & 52.2 & 45.7 & 49.4 \\
		
		UniSeg (Null BCE) & & 56.1 & 46.2 & \textbf{52.3} & \textbf{48.1} & 50.7 \\
		
		UniSeg (C-R BCE) & & \textbf{56.8} & \textbf{47.6} & 52.1 & \textbf{48.1} & \textbf{51.2}\\
		
		\midrule
		Multi-dataset (CE) & \multirow{3}{*}{\shortstack{HRNet\\W48}} & 62.1 & 49.2 & 56.8 & 51.8 & 55.0 \\
		
		UniSeg (Null BCE) & & 64.6 & \textbf{53.3} & \textbf{58.1} & \textbf{53.4} & 57.4 \\
		
		UniSeg (C-R BCE) & & \textbf{65.8} & 52.9 & 57.9 & \textbf{53.4} & \textbf{57.5} \\
		
		\bottomrule
	\end{tabular}}
	\vspace{-4mm}
\end{table}

\subsection{Results on WildDash2 Benchmark}

We further highlight the effectiveness of UniSeg by evaluating on the WildDash2 (WD2) benchmark. The WD2 benchmark is a new version of the original WildDash dataset, with a few additional classes and negative samples.
To evaluate on the WD2 benchmark, we employ the HRNet-W48 ``Multi-dataset'' and UniSeg models trained on all four datasets (C + I + B + M setting of Table~\ref{table:hrnet_big}).
Only the test images are provided to users, while evaluation is done on the WD2 server.

Our UniSeg model currently sits at the fourth place on the public leaderboard\footnote[1]{\url{https://wilddash.cc/benchmark/summary_tbl?hc=semantic_rob_2020}}, only surpassed by a method with a more powerful architecture that uses WD2 as the training set, while we do not use any data in WD2 during training. A summary of the results in shown in Table~\ref{table:wd2}.
Note that, while MSeg~\cite{mseg} merges some important fine-grained classes such as ``road markings'', our UniSeg model is able to make predictions for such classes. Also, a recent concurrent work \cite{yin2022devil} replaces class labels with text descriptions for achieving multi-dataset training, where they add open datasets \cite{OpenImages, Objects365} to training.

\begin{table}[!h]
	\caption{
		Class mIoU and negative class mIoU on the WildDash2 benchmark.
	}
	\vspace{-2mm}
	\label{table:wd2}
	\small
	\centering
	\renewcommand{\arraystretch}{1.1}
	\setlength{\tabcolsep}{2.5pt}
	\resizebox{0.47\textwidth}{!}{
	\begin{tabular}{llccc}
		\toprule
		Method & Architecture & Class mIoU & Negative mIoU & Meta Avg.  \\
		\midrule
		MSeg~\cite{mseg} & HRNet-W48 & 38.7 & 24.7 & 35.2 \\
		Yin \textit{et al.}~\cite{yin2022devil} & HRNet-W48 & - & - & 35.7 \\
		Yin \textit{et al.}~\cite{yin2022devil} & Segformer-B5& - & - & 37.9 \\
		\midrule
		Multi-Dataset (CE) & HRNet-W48 &39.0 & 27.9 & 36.0 \\
		UniSeg (C-R BCE) & HRNet-W48 & \textbf{41.7} & \textbf{34.8} & \textbf{39.4} \\
		\bottomrule
	\end{tabular}}
\end{table}

\subsection{Qualitative Analysis}

To better understand the full capacity of our UniSeg model, we visualize the output predictions of the UniSeg (C-R BCE) and CE models on two samples of the Cityscapes validation set in Figure~\ref{fig:visualization}. 
We first normalize the logits for each model's output, which is done by computing the softmax across all 70 classes for the CE model, and an element-wise sigmoid operation for the UniSeg model. 
The classes of Cityscapes with the top-1 scores are plotted to obtain the predictions in column 2. Next, we identify the top-1 classes among the non-Cityscapes classes and plot the scores in column 3. Finally, we obtain multi-label predictions by thresholding the scores of the non-Cityscapes classes: if the score is above a set threshold, the original class is replaced with the non-Cityscapes class. Here, we use 0.5 as the threshold for the UniSeg model, and 0.1 for the CE model. 

Through this visualization, we observe that our model exhibits interesting properties beyond the quantitative results. First, we find that our model can \textbf{make accurate predictions even for regions where Cityscapes does not provide a ground truth}: in the first sample, although the backside of the traffic signs are not labeled (black in ground truth) in Cityscapes, our model outputs high scores for these pixels (column 3) and overrides the original prediction to the ``traffic sign - back'' class from the Mapillary dataset (column 4). Furthermore, in the second sample, we observe similar behavior with the ``street lights'' class (beige). 

Our model also \textbf{effectively handles cases with direct label conflict}. In the first sample we see men on a motorcycle, which is given the ``rider'' label in Cityscapes (column 2). However, since our model is also trained on the ``motorcyclist'' class, and is able to alleviate the gradient conflict between ``rider'' and ``motorcyclist'', our model generates large activations for ``motorcyclist'' (light purple in column 4) as well. Similar results can be seen for the second sample, where the ``lane marking'' class (white) replaces parts of the ``road'' class, and the ``bicyclist'' class (brown) overrides the ``rider'' class. Note that, unlike the UniSeg model, the model trained on CE cannot produce large activations for these conflicting labels.

\section{Conclusions}
\vspace{-1mm}
In this paper, we propose UniSeg, which is an effective method to train multi-dataset segmentation models with different label spaces. To alleviate the gradient conflict issue caused by conflicting labels across datasets, we designed a ``Null'' class strategy using the class-independent BCE loss. To further reap the benefits of multi-dataset training, we learn class relationships and incorporate it into a new class-relational BCE loss.
In experiments, our model demonstrates improvements over the ordinary multi-dataset training, especially for the unseen datasets. We hope that the proposed method will act as a stepping stone for further development in training multi-dataset semantic segmentation models.
%

{\small
\bibliographystyle{ieee_fullname}
\bibliography{egbib}
}

\clearpage
\appendix
\onecolumn


\setcounter{figure}{0}
\setcounter{table}{0}
\setcounter{section}{0}

\renewcommand{\thepage}{A-\arabic{page}}
\renewcommand{\thesection}{Appendix \Alph{section}}
\renewcommand{\thetable}{A-\arabic{table}}
\renewcommand{\thefigure}{A-\arabic{figure}}
\renewcommand{\theequation}{a-\arabic{equation}}

\section{More Examples of Multi-Label Predictions}
\begin{figure*}[h!]
    \begin{tabular}{cccc}
        \multicolumn{4}{c}{\includegraphics[width=\textwidth]{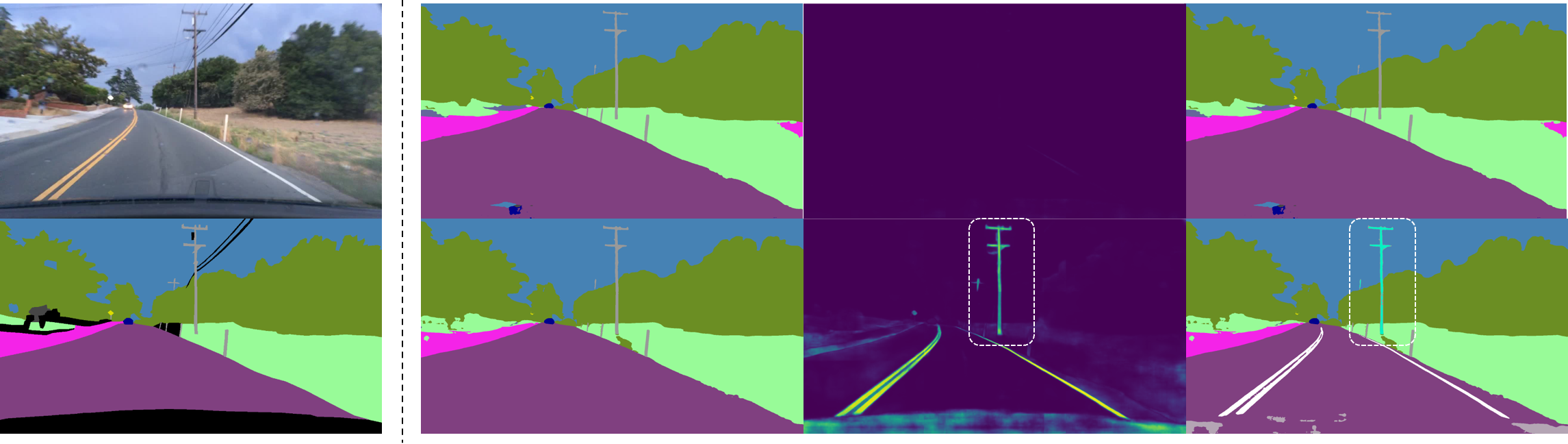}}\\
        \hspace{1cm}\shortstack{\small{Input image} \& \\ \small{ground truth}} 
        & \hspace{2.5cm}\shortstack{\small{Prediction on} \\ \small{BDD label space}} 
        & \hspace{1cm}\shortstack{\small{Normalized top1 activation} \\ \small{for non-BDD classes}} 
        & \shortstack{\small{Multi-label prediction:} \\ \small{entire label space}}
    \end{tabular}
    \vspace{-0.3cm}
    \caption{Multi-class predictions of the \textbf{BDD} dataset. The first row corresponds to an HRNet-W48 model trained with the CE loss, while the second row corresponds to our C-R BCE model. While both models make strong predictions on the BDD label space (column 2), only the C-R BCE model has high (normalized) activations for non-BDD classes in regions with label conflict (column 3). For example, Lane markings (color: white), while not labeled in BDD, are predicted via multi-class prediction, as well as the ``utility pole'' class (color: green) from the Mapillary dataset.}
    \vspace{-0.5cm}
    \label{fig:bdd_multiclass}
\end{figure*}

\begin{figure*}[h!]
    \begin{tabular}{cccc}
        \multicolumn{4}{c}{\includegraphics[width=\textwidth]{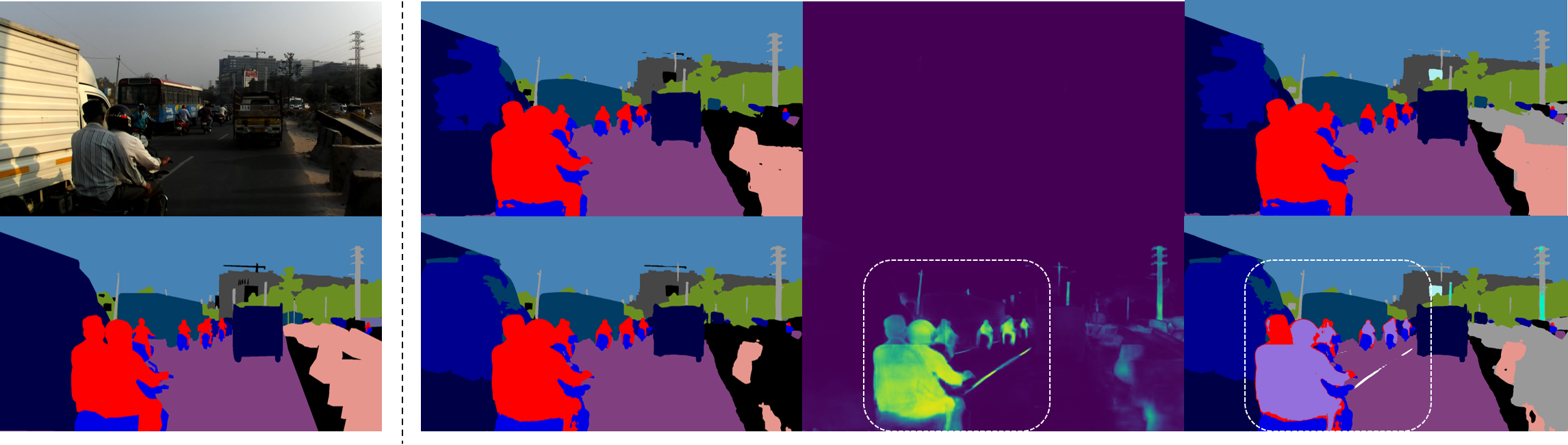}}\\
        \hspace{1cm}\shortstack{\small{Input image} \& \\ \small{ground truth}} 
        & \hspace{2.5cm}\shortstack{\small{Prediction on} \\ \small{IDD label space}} 
        & \hspace{1cm}\shortstack{\small{Normalized top1 activation} \\ \small{for non-IDD classes}} 
        & \shortstack{\small{Multi-label prediction:} \\ \small{entire label space}}
    \end{tabular}
    \vspace{-0.3cm}
    \caption{Multi-class predictions of the \textbf{IDD} dataset. Most ``riders'' are predicted as ``motorcyclists'' (color: light purple) in the multi-class prediction.}
    \label{fig:idd_multiclass}
\end{figure*}

\begin{figure*}[h!]
    \begin{tabular}{cccc}
        \multicolumn{4}{c}{\includegraphics[width=\textwidth]{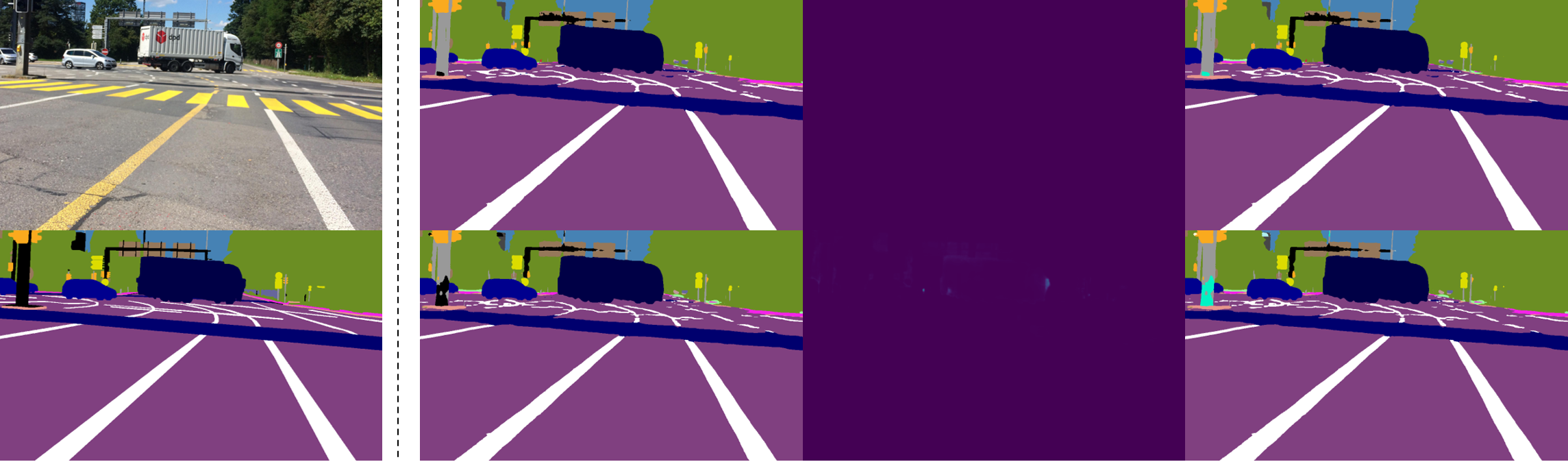}}\\
        \hspace{1cm}\shortstack{\small{Input image} \& \\ \small{ground truth}} 
        & \hspace{2cm}\shortstack{\small{Prediction on} \\ \small{Mapillary label space}} 
        & \hspace{1cm}\shortstack{\small{Normalized top1 activation} \\ \small{for non-Mapillary classes}} 
        & \shortstack{\small{Multi-label prediction:} \\ \small{entire label space}}
    \end{tabular}
    \caption{Multi-class predictions of the \textbf{Mapillary} dataset. There is almost no noticeable difference between the predictions of the CE and C-R BCE model. This figure shows that the CE model is capable of predicting fine-grained labels. However, it will not do so for domains that did not have these fine-grained labels in the first place.}
    \label{fig:mapillary_multiclass}
\end{figure*}

In Figures~\ref{fig:bdd_multiclass},~\ref{fig:idd_multiclass}, and~\ref{fig:mapillary_multiclass}, we present additional qualitative results from BDD, IDD, and Mapillary datasets, respectively.
These figures extend Figure 4 from the main paper, and aim to show that similar behavior can be observed in BDD and IDD.

\vspace{-0.3cm}

Interestingly, on Mapillary we notice that even the CE model can output strong activations for ``lane marking'', which was not the case when evaluating on datasets that do not label this class (Cityscapes: Figure 4, BDD: Figure~\ref{fig:bdd_multiclass}, and IDD: Figure~\ref{fig:idd_multiclass}).
Based on this observation, we argue that the CE model insidiously learns a connection between the domain and the label space.
This could be seen as a form of overfitting.
On the contrary, the C-R BCE model evidently does a much better job at generalizing the label space across domains.

\section{Predicted Multi-Labels}
\begin{table*}[h]
	\caption{Predicted multiple labels for Class-Relational BCE}
	\vspace{1mm}
	\label{table:multi_labels}
	\small
	\centering
	\begin{tabular}{l|r}
		\toprule
		Primary Label & Secondary Label \\
		\midrule
		bike lane & road \\
		catch basin & road \\
		crosswalk - plain & road \\
		lane marking - general & road \\
		parking & road \\
		pothole & road \\
		service lane & road \\
		junction box & obs-str-bar-fallback \\
		mailbox & obs-str-bar-fallback \\
		phone booth & obs-str-bar-fallback \\
		traffic sign (back) & obs-str-bar-fallback \\
		trash can & obs-str-bar-fallback\\
		curb cut & sidewalk \\
		pedestrian area & sidewalk \\
		bicyclist & rider \\
		motorcyclist & rider \\
		ground animal & person \\
		other rider & person \\
		banner & billboard \\
		phone booth & building \\
		caravan & car \\
		on rails & train \\
		trailer & truck \\
		\bottomrule
	\end{tabular}
\end{table*}
Table~\ref{table:multi_labels} presents the list of multi-labels that are predicted by our model and used by the Class-Relational BCE loss.
The ``Primary Label'' column lists the names of the original categories, while the ``Secondary Label'' column lists the corresponding multi-label, \ie a pixel labeled as ``bike lane'' will also receive supervision for the ``road'' class.
As mentioned in the paper, these relationships are not symmetric, \ie a pixel of ``road'' will not necessarily lead to supervision of ``bike lane''.

The list of primary and secondary labels in Table~\ref{table:multi_labels} supports our hypothesis from the main paper.
As expected, the ``bicyclist'' and ``motorcyclist'' classes are both mapped to ``rider'' as well, which means that any supervision on those classes will also help learn stronger representations for the ``rider'' class. In the CE setting, the model would only receive negative supervision (label of 0) for the ``rider'' class from Mapillary data, and in the Null BCE setting, the model would receive no supervision for the ``rider'' class from Mapillary data.
However, in our C-R BCE setting, Mapillary data of the ``bicyclist'' and ``motorcyclist'' classes would also provide positive supervision for the ``rider'' class, so the model can learn a more robust representations for ``rider''.

\section{Importance of our BCE Loss}
In the main paper, we outline the issues encountered when using the cross-entropy (CE) loss function, and propose to use the binary cross-entropy (BCE) loss as a replacement to alleviate these issues.
Here we further discuss the importance of the BCE loss in our method and how tightly this new loss function is coupled with the second-stage training using class-relational BCE loss.

In Section 3.3 of the main paper, the class relationship is more meaningful when the underlying model is trained with the BCE loss. The reason is that, BCE loss operates on a per-class basis, and thus the output scores can be high for multiple classes. This property is useful when using class relationships as some categories may correlate to each other during model training.
On the contrary, using the CE loss based on Softmax only outputs the probability of a certain class, describing how probable that class is the ground-truth, \textit{relative to all other classes}, and thus it cannot exploit the class relationships as the BCE loss does.

Furthermore, the BCE loss enables us to train with multiple ground-truth classes on a single pixel. This property is especially useful when training with multiple datasets that have different label spaces, since classes may not be fully disjoint to each other, \ie they may actually refer to the same class or one class may be a subset of another. Examples of such cases are described under the ``Discussions'' paragraph in Section 3.3.

\section{Results on Seen Datasets}

\begin{table}[!h]
	\caption{
		mIoU comparisons for the seen datasets in the C-I-B-M setting with HRNet-W18 and HRNet-W48.
	}
	\label{table:seen_datasets}
	\centering
	\begin{tabular}{llccccc}
		\toprule
		Method & Arch. & Cityscapes & IDD & BDD & Mapillary & Mean \\
		\midrule
		Multi-dataset (CE) & \multirow{3}{*}{\shortstack{HRNet\\W18}} & 71.8 & 63.1 & 60.4 & 37.2 & 58.1 \\

		UniSeg (Null BCE) & & 71.7 & 63.2 & 60.2 & 36.1 & 57.8  \\

		UniSeg (C-R BCE) & & 71.5 & 63.0 & 60.0 & 35.4 & 57.5 \\

		\midrule
		Multi-dataset (CE) & \multirow{3}{*}{\shortstack{HRNet\\W48}} & 80.4 & 70.4 & 67.1 & 46.4 & 66.1 \\

		UniSeg (Null BCE) & & 80.9 & 71.1 & 68.0 & 45.5 & 66.4  \\

		UniSeg (C-R BCE) & & 81.0 & 70.9 & 67.4 & 43.0 & 65.6 \\

		\bottomrule
	\end{tabular}
\end{table}

We present the results of the seen datasets for the ``C-I-B-M (All)'' setting in Table~\ref{table:seen_datasets}. As seen in this table, our Null BCE and C-R BCE models maintain competitive performance to the CE baseline even when evaluating on the seen datasets of their original label space in individual datasets, but not on the unified label space from all the datasets. Note that, for datasets like Mapillary that contain fine-grained categories, the performance drops more as the CE model pays more attention to fine-grained information. However, the benefit of our model is orthogonal to training a good fine-grained model, since our goal is to train multiple datasets together and generalize to unseen datasets in the unified label space.

%

\newpage

\section{Qualitative Results on Unseen Datasets}
\begin{figure*}[h!]
    \includegraphics[width=\textwidth]{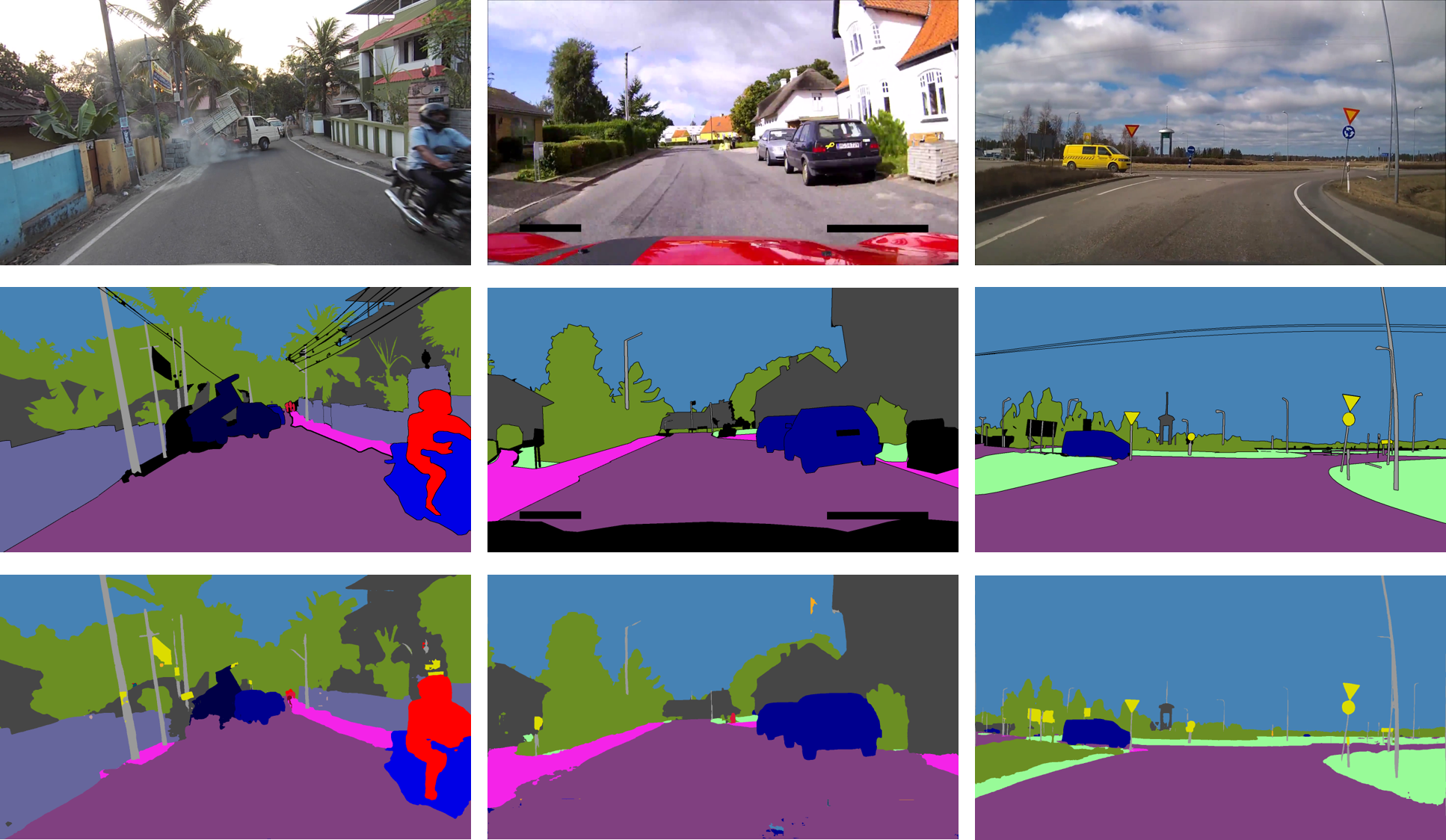}
    \caption{Additional WildDash visualizations. From top to bottom: Original image, Ground Truth, UniSeg model prediction.}
    \label{fig:supp_wilddash}
\end{figure*}

\begin{figure*}[h!]
    \includegraphics[width=\textwidth]{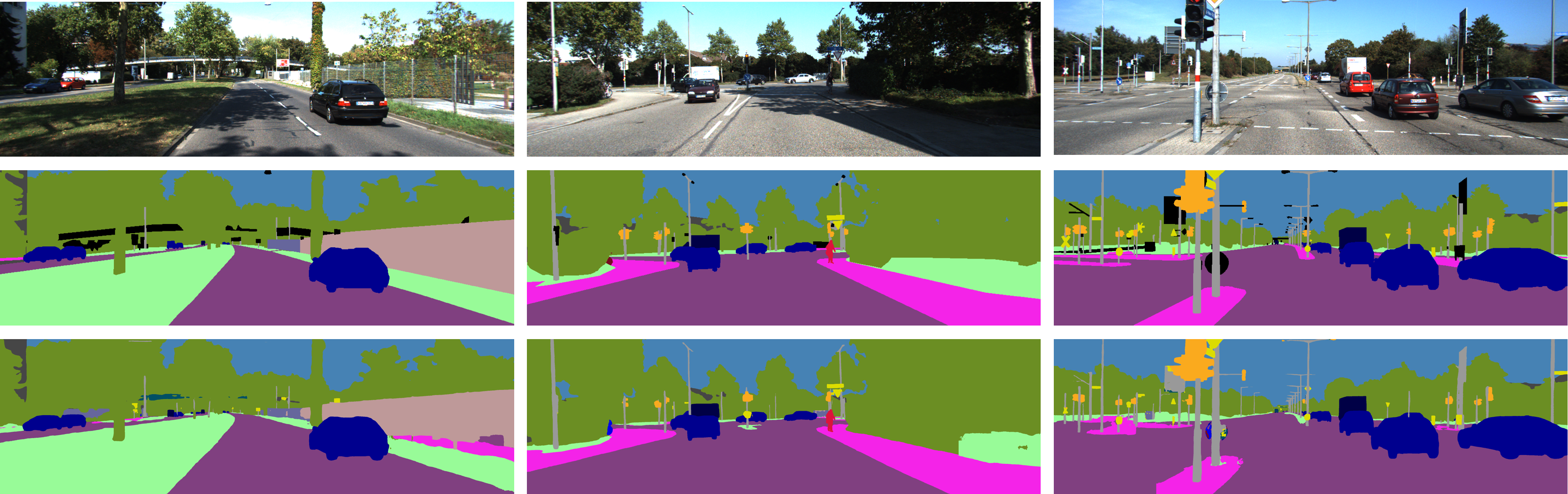}
    \caption{Additional KITTI visualizations. From top to bottom: Original image, Ground Truth, UniSeg model prediction.}
    \label{fig:supp_kitti}
\end{figure*}

\begin{figure*}[h!]
    \includegraphics[width=\textwidth]{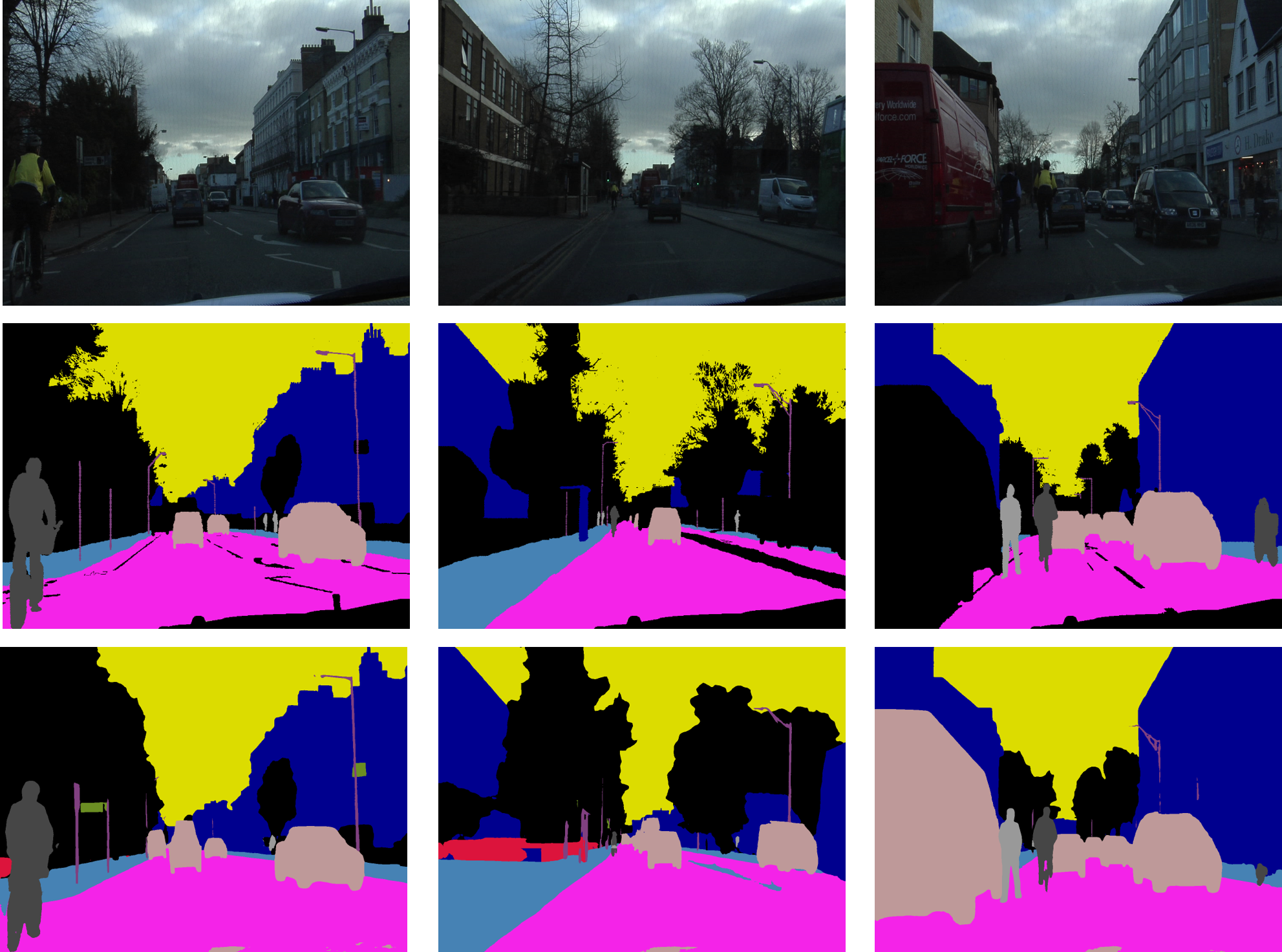}
    \caption{Additional CamVid visualizations. From top to bottom: Original image, Ground Truth, UniSeg model prediction.}
    \label{fig:supp_camvid}
\end{figure*}

In Figure \ref{fig:supp_wilddash}, \ref{fig:supp_kitti}, \ref{fig:supp_camvid}, we show qualitative results for semantic segmentation on the unseen datasets, \ie, WildDash, KITTI, and CamVid. Note that, although these input images are not from the training datasets, our model is still able to provide accurate segmentation predictions compared to the ground truths.
All predictions are made on the label space of the input image.

\clearpage
\newpage
\section{Derivation of Gradients}

We derive the gradients for the cross-entropy loss (Eq. (2) in the main paper) and the binary cross entropy loss (Eq. (4) in the main paper).
First, let us simplify our formulation and refer to just a single output location, $(h, w)$, of the segmentation map from the $i^{\text{th}}$ dataset.
\begin{equation}
    \mathcal{L}_{seg}^{ce} = - \sum_{k=1}^{K_u} Y^{(k)} \log(P^{(k)}).
\end{equation}
Given an ground truth $y$, the one-hot label $Y^{(k)} = 1$ when $k = y$ and $Y^{(k)} = 0$ when $k \neq y$. Thus, this equation can be simplified as:
\begin{equation}
    \mathcal{L}_{seg}^{ce} = - \log(P^{(y)}).
\end{equation}
With the cross-entropy loss, we use a softmax function over the channel dimension on the output, $O$. Hence, substituting $P^{(y)}$ with the softmax over $O$ obtains:
\begin{equation}
    \mathcal{L}_{seg}^{ce} = - \log \left( \frac{e^{O^{(y)}}}{\sum_{k} e^{O^{(k)}}} \right) = \log \left(\sum_{k} e^{O^{(k)}}\right) - O^{(y)}.
\end{equation}
Now, taking the gradient of $\mathcal{L}_{seg}^{ce}$ with respect to the output of an arbitrary parameter $\theta$:
\begin{align}
    \label{eq:chain}
       \frac{\partial \mathcal{L}_{seg}^{ce}}{\partial \theta} &= \sum_{k} \frac{\partial \mathcal{L}_{seg}^{ce}}{\partial O^{(k)}} \frac{\partial O^{(k)}}{\partial \theta}.
\end{align}
In the paper, we leave $\frac{\partial O^{(k)}}{\partial \theta}$ as is, since it is irrelevant for this derivation. Instead, we focus on the $\frac{\partial \mathcal{L}_{seg}^{ce}}{\partial O^{(k)}}$ term.

\begin{align}
\label{eq:supp_derivative_ce}
    \frac{\partial \mathcal{L}_{seg}^{ce}}{\partial O^{(k)}} &= \frac{\partial}{\partial O^{(k)}}  \log \left(\sum_{k'} e^{O^{(k')}}\right) - \frac{\partial}{\partial O^{(k)}} O^{(y)} \nonumber \\
    &= \frac{O^{(k)}}{\sum_{k'} e^{O^{(k')}}} - \frac{\partial}{\partial O^{(k)}} O^{(y)} \nonumber \\
    &= \frac{O^{(k)}}{\sum_{k'} e^{O^{(k')}}} - Y^{(k)} \nonumber \\
    &= P^{(k)} - Y^{(k)},
\end{align}
As seen in Eq.~\eqref{eq:supp_derivative_ce}, when $k$ points to a conflicting class in which two datasets provide different labels, we end up with the gradient conflict depicted in Eq. (2) of the main paper.
Using a similar process with the aforementioned simplifications for the binary cross-entropy loss:
\begin{equation}
    \label{eq:supp_bce}
    \mathcal{L}_{seg}^{bce} = - \sum_{k}^{K_i} Y^{(k)} \log(Q^{(k)}) + (1 - Y^{(k)}) \log(1 - Q^{(k)})),
\end{equation}
where $Q$ denotes the sigmoid-activated output, $O$. Since
\begin{equation}
    \frac{\partial Q^{(k)}}{\partial O^{(k)}} = Q^{(k)}(1 - Q^{(k)}),
\end{equation}
we calculate the gradient of the BCE loss with respect to an output, $O^{(c)}$:

\begin{align}
    \frac{\partial \mathcal{L}_{seg}^{ce}}{\partial O^{(c)}} &= - \frac{Y^{(c)}}{Q^{(c)}} Q^{(c)}(1 - Q^{(c)}) + \frac{(1 - Y^{(c)})}{1 - Q^{(c)}} Q^{(c)}(1 - Q^{(c)}) \nonumber \\
    &= Y^{(c)} (Q^{(c)} - 1) +  Q^{(c)}(1 - Y^{(c)}).
    \label{eq:supp_bce_grad}
\end{align}
Again, depending on the ground truth class, the righthand term in Eq.~\eqref{eq:supp_bce_grad} simplifies to either $Q^{(c)} - Y^{(c)}$ or $Y^{(c)} - Q^{(c)}$.

In Eq. (2) of the main paper, we omitted the sum over $k$ for the gradient of the cross-entropy loss with respect to parameter $\theta$. The fix is reflected in~\eqref{eq:chain}. Note that this change does not affect the conclusion we make from these equations.

\section{Cosine Classifier}
We replace the final classification layer in the segmentation model with a cosine classifier~\cite{cosface}.
It is implemented via $\ell_2$-normalization of both the $1 \times 1$ convolution weights and extracted features across the channel dimension. Let $\mathbf{\hat{\phi_c}}$ denote the $\ell_2$-normalized $1 \times 1$ convolution weight vector for the $c$\textsuperscript{th} class, and $\hat{\mathbf{x}}^{(h,w)}$ denote the $\ell_2$-normalized input feature vector at location $(h, w)$. Then, the cosine similarity for class $c$ at location $(h, w)$ is calculated as:
\begin{align}
    \label{eq:cosine}
    S^{(h, w, c)} = t \cdot \mathbf{\hat{\phi_c}}^\top \hat{\mathbf{x}}^{(h,w)} = t \cdot ||\mathbf{\phi_c}|| ||\mathbf{x}^{(h,w)}|| \cos \theta_c,
\end{align}
where $\theta_c$ represents the angle between $\phi_c$ and $\mathbf{x}^{(h,w)}$, and $t$ is a scaling factor (we use $t = 20$ in the paper).

\section{Union of Categories}

\rowcolors{2}{gray!25}{white}
\begin{table*}[h]
    \caption{Unified label space for Cityscapes, BDD, IDD, and Mapillary: there are 70 categories and we list them for individual datasets.}
    \vspace{1mm}
    \label{table:category_union}
    \small
    \centering
	\resizebox{0.95\textwidth}{!}{
    \begin{tabular}{cccc||cccc}
        \toprule
        \textbf{Cityscapes} & \textbf{BDD} & \textbf{IDD} & \textbf{Mapillary} & \textbf{Cityscapes}   & \textbf{BDD}  & \textbf{IDD}      & \textbf{Mapillary}   \\
        \midrule
        & & autorickshaw & & & & obs-str-bar-fallback & \\
        
        & & & banner & & & & on rails \\ 
        &              &              & barrier              &                     &               &                  & other rider       \\ 
        &      &      & bench            &                     &               &          & other vehicle             \\ 
        bicycle                    & bicycle              & bicycle              & bicycle          &                     &               & parking                  & parking     \\ 
        &              &              & bicyclist          &              &        &           & pedestrian area              \\ 
        &              &              & bike lane          & person                     & person               & person                  & person         \\ 
        &              &    & bike rack          &                &          &             & phone booth                \\ 
        &              & billboard              & billboard               & pole                     & pole               & pole                  & pole             \\ 
        &              &              & bird               &                     &               &       & pothole          \\ 
        &              &       & boat             &               &         & rail track            & rail track                     \\ 
        &     & bridge     & bridge           & rider                & rider          & rider             &                \\ 
        building                 & building          & building          & building                & road                     & road               & road                  & road                \\ 
        bus                 & bus          & bus          & bus                &                     &               &                  & sand        \\ 
        car                    & car              & car              & car          &            &      &         & service lane            \\ 
        &              &              & car mount            & sidewalk                 & sidewalk           & sidewalk              & sidewalk                 \\ 
        &              &              & caravan        & sky                     & sky               & sky                  & sky                \\ 
        &              &              & catch basin        &                     &               &                  & snow        \\ 
        &              &              & cctv camera  &             &       &                  & street light             \\ 
        &              &         & crosswalk - plain               & terrain       & terrain &    & terrain       \\ 
        &              & curb              & curb           & traffic light        & traffic light  & traffic light     & traffic light        \\ 
        &              &              & curb cut        & traffic sign                     & traffic sign               & traffic sign                  & traffic sign \\ 
        &        &        & ego vehicle              &                     &               &                  & traffic sign (back)  \\ 
        fence                    & fence              & fence              & fence       &                     &               &                  & traffic sign frame             \\ 
        &              &              & fire hydrant      &               &         &                  & trailer                     \\ 
        &              &   & ground animal         & train                     & train               &                  &           \\ 
        &              & guard rail              & guard rail       &               &         &            & trash can               \\ 
        & & & junction box & truck & truck & truck & truck \\ 
        & & & lane marking - crosswalk          &              &              & tunnel            \\ 
        &              &              & lane marking - general          &   &   & utility pole  &       \\
        &              &              & mailbox       & vegetation                     & vegetation               & vegetation                  & vegetation               \\ 
        &              &              & manhole           &                     &               & vehicle fallback                  &        \\ 
        motorcycle                    & motorcycle              & motorcycle              & motorcycle           & wall                     & wall               & wall                  & wall        \\ 
        &              &              & motorcyclist           &                     &               &                  & water        \\ 
        &              &              & mountain           &                     &               &                  & wheeled slow        \\
        \bottomrule
    \end{tabular}
    }
\end{table*}

\section{Potential Negative Impact and Limitations}
One potential negative impact of our work is that it could be applied in surveillance systems, which are always a topic of controversy. To mitigate the unintended use of our work for unlawful surveillance, our downstream applications will be accompanied with precautions to highlight this risk.
Also, there could be privacy concerns with data collection, although most segmentation datasets do a good job of removing any identifiable information from the images. 

A limitation of our work is that by mixing fine-grained labels with coarser ones, it may be harder for the model to predict the fine-grained classes compared to the model that is trained on just the fine-grained dataset.
However, we argue that in our case the benefits still outweight the costs --- by training with multiple datasets, the model is able to better generalize to new domains.
In fact, as shown in Figure 4 the main paper, our Null BCE and C-R BCE loss functions enable the model to make multi-label predictions for fine-grained classes.
\end{document}